\documentclass{article}
\usepackage{spconf,amsmath,graphicx}
\usepackage{bbold}
\usepackage{makecell}
\usepackage{graphics}
\usepackage{multirow}
\usepackage{tabularray}
\usepackage{bm}
\usepackage{hyperref}
\usepackage{makecell}
\usepackage{float}
\usepackage{arydshln}
\usepackage{multirow}
\usepackage{pifont}% http://ctan.org/pkg/pifont
\usepackage{amssymb}
\usepackage{subcaption}
\usepackage{soul}
\usepackage{graphicx} % omit 'demo' for real document
\usepackage[dvipsnames]{xcolor}

\usepackage{url}
\usepackage{multirow}
\usepackage{graphicx}
\usepackage{tabularx}
\newcommand{\uls}[1]{\underline{\smash{}}}

\usepackage{enumitem}
\setlist{nosep, leftmargin=14pt}

\usepackage{mwe} % to get dummy images

\title{Guiding the classification of hepatocellular carcinoma on 3D CT-scans using deep and handcrafted radiological features}
\name{
E. Sarfati$^{1,2}$ \qquad A. Bône$^{1}$ \qquad M-M. Rohé$^{1}$ \qquad C. Aubé$^{4}$ \qquad M. Ronot$^{5,6}$ \qquad  P. Gori$^{2}$ \qquad I. Bloch$^{2,3}$}
\address{$^{1}$ Guerbet Research, Villepinte, France \\
 $^{2}$ LTCI, Télécom Paris, Institut Polytechnique de Paris, Saclay, France \\
 $^{3}$ Sorbonne Université, CNRS, LIP6, Paris, France\\
 $^{4}$ Département de Radiologie, Centre Hospitalier Universitaire d’Angers, Angers, France \\ 
 $^{5}$ Service de Radiologie, HUPNSV, Hôpital Beaujon, Clichy, France\\
 $^{6}$ INSERM UMR 1149, Université de Paris, Paris, France}
\begin{document}
\ninept

\maketitle

\begin{abstract}
Hepatocellular carcinoma is the most spread primary liver cancer across the world ($\sim$80\% of the liver tumors). The gold standard for HCC diagnosis is liver biopsy. However, in the clinical routine, expert radiologists provide a visual diagnosis by interpreting hepatic CT-scans according to a standardized protocol, the LI-RADS, which uses five radiological criteria with an associated decision tree. In this paper, we propose an automatic approach to predict histology-proven HCC from CT images in order to reduce radiologists' inter-variability. We first show that standard deep learning methods fail to accurately predict HCC from CT-scans on a challenging database, and propose a two-step approach inspired by the LI-RADS system to improve the performance. We achieve improvements from 6 to 18 points of AUC with respect to deep learning baselines trained with different architectures. We also provide clinical validation of our method, achieving results that outperform non-expert radiologists and are on par with expert ones.
\end{abstract}

\section{Introduction}

Liver cancer is the sixth  most common cancer worldwide and represents the fourth cause of mortality in 2023 \cite{livercancer}. In particular, hepatocellular carcinoma (HCC) accounts for 80\% of the primary liver cancers \cite{livercancer,hcc}. Diagnosing the presence of hepatocellular carcinoma is obtained with a biopsy, which is a risky and invasive operation. On the other hand, radiological assessment can provide a non-invasive diagnosis, based on typical features on dynamic CT or MRI according to strict criteria described in international guidelines \cite{hccmri,ying,shi}. This procedure is considered a solid proxy of the presence or absence of the tumor, and often totally replaces liver biopsy in clinical routine. %At this aim, 
To perform this assessment, radiologists use a standardized scoring system called LI-RADS (Liver Imaging Reporting and Data System)~\cite{lirads} to characterize suspicious lesions on CT-scans. The latter uses the four phases of the traditional injection protocol for CT imaging, and is based on five main visual features. However, radiological diagnosis suffers from an inherent variability between experts, which can be reduced using automatic methods. Several methods have been proposed in the literature to improve the diagnosis of hepatocellular carcinoma using deep learning \cite{shi,ling,bone,ronot,wang,sarfatiecr}. The authors in \cite{bone,yingnature} predict directly the pathology - without using any intermediate features - but exploit very large medical private datasets~\cite{yingnature} ($\sim$12,000 patients in their study) that present a huge size difference between non-HCC and HCC lesions, which is a well-known feature to characterize HCC but can induce a bias in the problem setting as their HCC lesion can be considered ``easy" or unuseful in the clinical process, due to the advanced stage of the tumor. In this work, we propose to evaluate an automated HCC diagnosis method on a challenging dataset of patients with risk factors of HCC and small-size nodules. As a baseline approach, we train deep learning baseline models inspired by state-of-the-art results proposed in the literature \cite{ling,yasaka}. To improve those results on our datasets, we propose a two-step approach, directly inspired by the radiologists' reading grid \cite{lirads}. This approach uses the LI-RADS radiological criteria for a preliminary learning task in order to improve the final prediction of HCC. To the best of our knowledge, no methods have been proposed in the literature to use the LI-RADS major features for histological HCC prediction on CT-scans. Our contributions are the followings:
\begin{itemize}
    \item[$\bullet$] We propose to improve the baseline classification of %histological 
    HCC using weak radiological labels (LI-RADS major features), used for a preliminary learning task.
    \item[$\bullet$] We propose new manually handcrafted features well-designed for HCC classification, as well as LI-RADS based deep learning radiological features, which can be combined with handcrafted features to enhance the classification performance. 
    \item[$\bullet$] We train and evaluate (cross-validation) our method on a challenging database of small and difficult HCC tumors \cite{ronot,aube}, and transfer it to one private test set. 
    \item[$\bullet$] Eventually, we compare our method to deep learning baseline methods and to liver expert and non-expert radiologists' LI-RADS diagnosis.
\end{itemize}

\section{Method}

\begin{figure}[h!]
    \centering
    \includegraphics[width=0.48\linewidth]{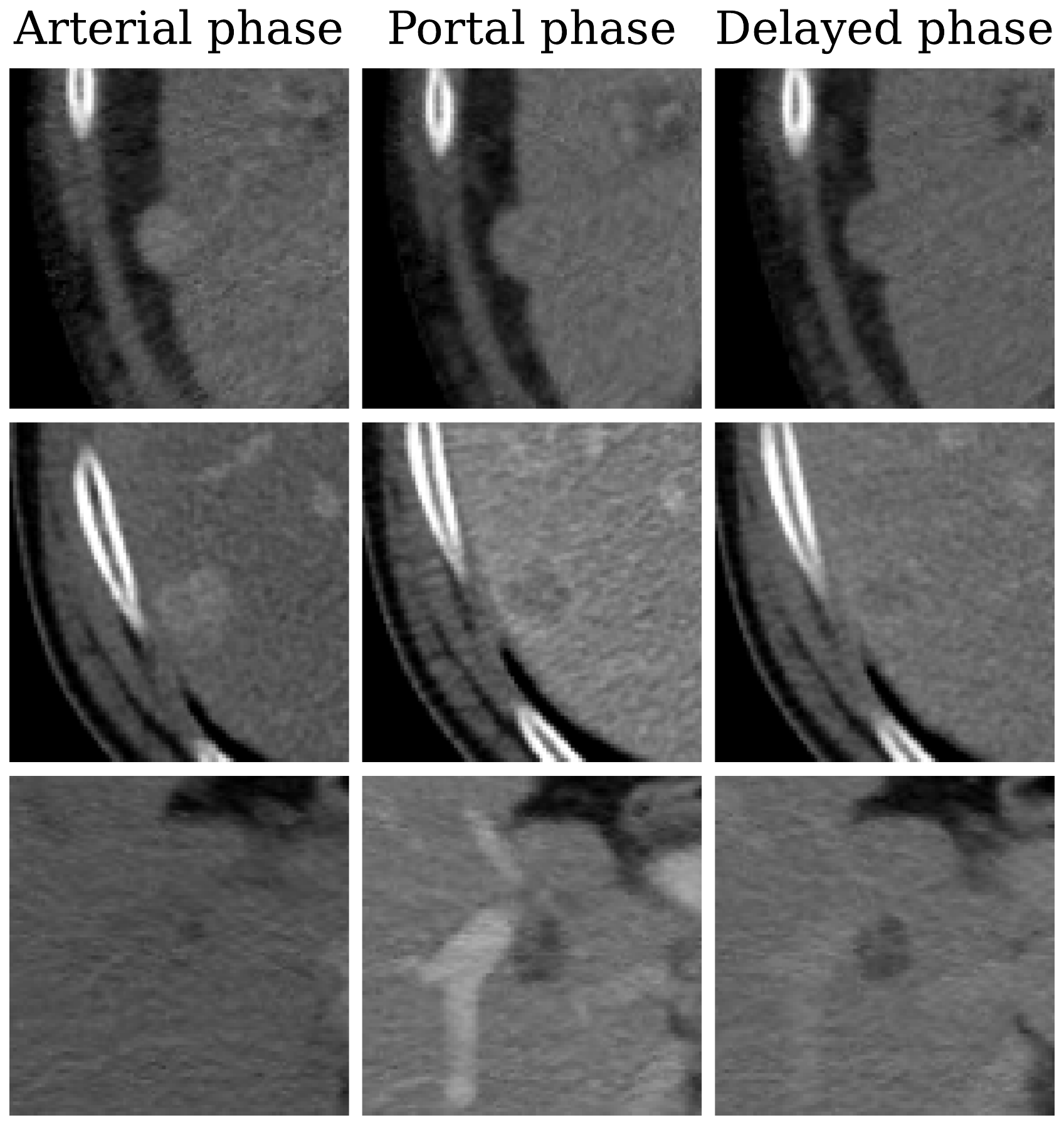}
    \caption{Examples of center slices extracted from $\mathcal{D}_1$.}
    \label{images}
\end{figure}

\textit{Deep learning features.} \quad
The proposed approach builds on a baseline, \textit{i.e.}. a simple supervised classifier using as ground-truth the histological confirmation of HCC, by adding expert knowledge. We propose to follow the LI-RADS scoring system decision tree \cite{ronot, wu,park} as a preliminary learning task. More precisely, we propose to combine the prediction of the three LI-RADS major features with the baseline HCC prediction in order to guide the final HCC classification. To do so, we introduce three deep learning sub-models, based on the same backbone architecture as the baseline one. We train our deep models on the three LI-RADS criteria separately, which gives us three probability outputs, one for each criterion. Furthermore, we add the size feature $s$, which can be automatically computed from the lesion segmentation.\\
\textit{Handcrafted features.} \quad
To increase the predictive power of the networks outputs, we propose to combine them with handcrafted radiomics features that are directly inspired of the radiologists' reading grid. 
\begin{figure}[ht]
    \centering
    \includegraphics[width=0.6\linewidth]{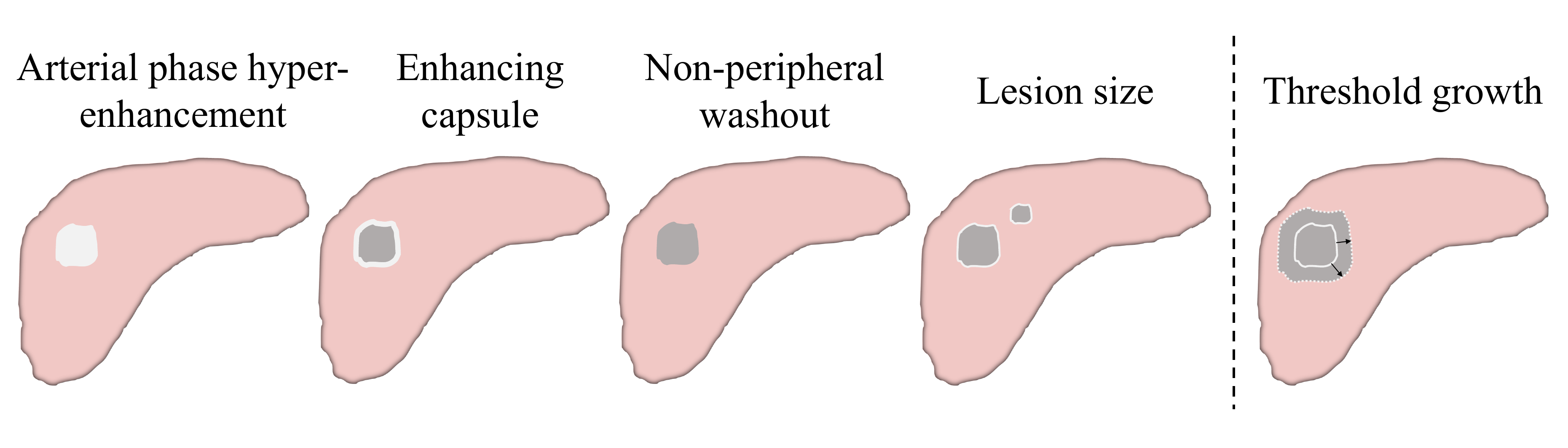}
    \caption{Schema of the LI-RADS five main visual features. The threshold growth, which is obtained with longitudinal scans, is not available in the datasets used in this study.}
    \label{schema}
\end{figure}
As represented in Figure~\ref{schema}, the LI-RADS binary criteria are visually assessed by observing the contrast differences of the lesions with respect to the liver parenchyma, or with the lesion borders. In this sense, it is intuitive to propose simple radiomics features computed on the input images. We propose three different formulas, based on the definition of each major feature \cite{majorfeatures}:
\begin{enumerate}
    \item[$\bullet$] \textbf{APHE}: the arterial phase hyper-enhancement is defined by a contrast difference between the lesion and the liver parenchyma. This contrast is observed on the arterial phase and can be seen on the portal venous phase, as shown in Figure~\ref{images}, lines one and two. 
    \item[$\bullet$] \textbf{EC}: the enhancing capsule is typically characterized by a thin enlightened contour with a darker inner lesion surface, as shown in Figure~\ref{images}, at the second and third lines. 
    \item[$\bullet$] \textbf{NPW}: the non-peripheral washout corresponds to a decrease in attenuation or intensity from earlier to later phase, resulting in hypoenhancement in the portal venous or delayed phase, \textit{e.g} a darkening of the lesion surface. It is observed at every line of Figure~\ref{images}.
\end{enumerate}

\begin{figure*}[ht]
    \centering
\includegraphics[width=0.6\linewidth]{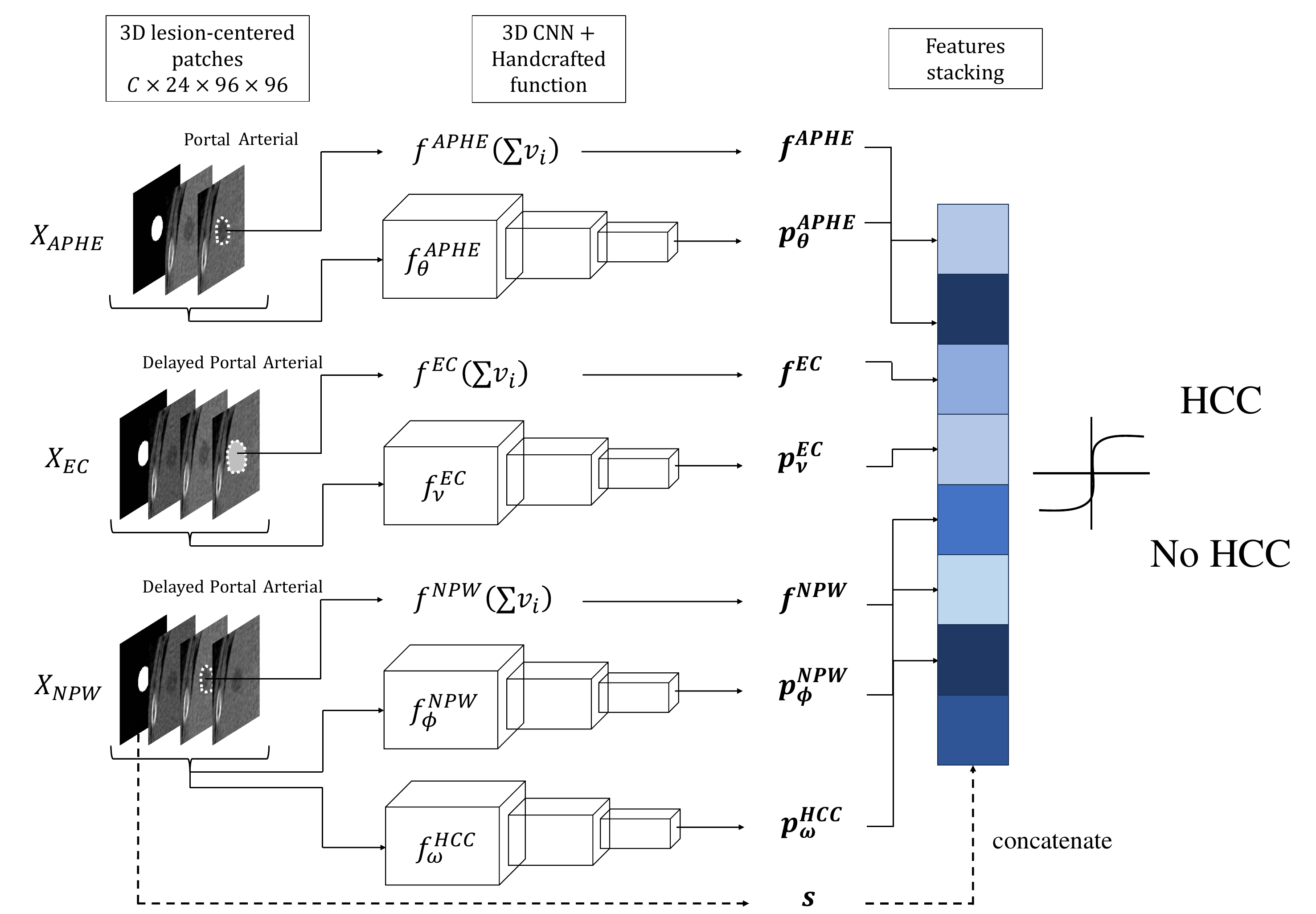}
    \caption{Overview of the proposed method. $s$ corresponds to the lesion diameter (in \textit{mm}). The architecture leads to a one-dimensional vector of 8 components, being either a handcrafted feature or a predicted probability of presence of each criterion.}
    \label{figmethod}
\end{figure*}

%\vspace*{5mm}
\noindent Following the previous criteria description, we propose to characterize the three features by three simple formulas. The APHE feature $f^{APHE}$ computes the difference of the median voxel values between the lesion and the parenchyma on the arterial phase. The same formulation is proposed for the NPW $f^{NPW}$, but on the portal venous phase. Finally, for the EC $f^{EC}$, we first propose to automatically compute the inner border of the lesion. Then we compute the difference of energy between the inner surface of the lesion and its border, which is a measure of the magnitude of voxel values in an image. Denoting $L$ the lesion, $P$ the liver parenchyma and $B$ the lesion border, we obtain:
\begin{align*}
f^{APHE} &= median\left(L_{arterial}\right) - median\left(P_{arterial}\right) \\
f^{EC} &= energy\left(L_{arterial}\right) - energy\left(B_{arterial}\right) \\
f^{NPW} &= median\left(L_{portal}\right) - median\left(P_{portal}\right)
\end{align*}
Naming $f$ the generic architecture used for the deep learning experiments, we denote by $f_\omega$, $f_\theta$, $f_\nu$ and $f_\phi$ the four distinct neural nets with parameters $\omega$, $\theta$, $\nu$ and $\phi$ and we propose to aggregate the outputs of all the models using a logistic regression giving us the final following model, with $\beta$ being the 7-dimensional vectors of the regression coefficients:
\begin{equation}
\begin{split}
    \hat{y}_{HCC} = \beta \left[p_\omega^{HCC},p_\theta^{APHE},p_\nu^{EC},p_\phi^{NPW},f^{APHE},f^{EC},f^{NPW} \right]^T \\ + \beta_8s + \alpha
\end{split}
\end{equation}
An overview of our method is presented in Figure~\ref{figmethod}.
\paragraph*{Image processing.} 
\noindent\textit{CT-scans processing.} \quad
We worked with original images of full CT-scans of dimensions $512\times512$ with anisotropic z-spacings. Images were first selected according to the presence of the three last injection phases: arterial phase, portal venous phase and delayed phase. First, a simple registration was made between all phases on the portal venous phase by performing a linear vertical translation (z-axis) based on the liver segmentations. Registered images were then resampled to match their portal venous phase scan using a nearest neighbor interpolation. To ensure homogeneity in all our dataset, we then used the nn-UNet pre-processing which is well adapted for z-anisotropic 3D images \cite{nnunet}, which allowed us to resample all our images to their median $0.76\times0.76\times2.00mm^3$ voxel size.
%resolution. \\

\noindent\textit{Patch sampling.} \quad
For lesion type classification, 3D patch inputs are commonly preferred \cite{bone,yasaka}, notably due to the limited computational resources which do not allow using the full CT-scans as inputs with a large enough batch size. Lesion-centered patches were sampled on the pre-processed 3D images so that 95\% of the segmented lesion fit in the boxes. Patches of size $96\times96\times24$ voxels were finally selected, corresponding to a $72.96\times72.96\times48.00mm^3$ field of view. According to the predicted major feature, the final inputs of our networks have a size of $(N,C,24,96,96)$ where $N$ is the batch size and $C$ the number of channels, with $C\in\{3,4\}$ if we use arterial/venous phases + lesion segmentation (APHE) or arterial/venous/delayed phases + lesion segmentation (HCC, EC and NPW).
\paragraph*{Datasets.} 
Two datasets were leveraged in this study, coming from different centers of acquisition. One was used for training and validation, the two other datasets were only used for testing purpose.\\
\noindent$\bm{\mathcal{D}_1}$. Our first dataset contains a total number of 244 lesions, coming from 182 distinct patients. Among these lesions, 161 are histologically-proven hepatocellular carcinoma and 83 are other types of lesion. This dataset is the only one of our datasets that contains the three LI-RADS radiological criteria, for each lesion. Each criterion is provided by a binary number corresponding to their presence or absence, evaluated by an expert radiologist. \\
\noindent$\bm{\mathcal{D}_2}$. \quad Our second dataset contains 1012 lesions corresponding to 543 patients and 602 no-HCC/410 HCC. This dataset presents cases of HCC that are mostly common \cite{yingnature,hcc_ct} as there exists a size gap between non-HCC and HCC lesions that we do not have in the first dataset. \\

One major feature for HCC characterization remains the size of the lesion \cite{hcc,lirads}. A larger lesion has indeed more chance of being an HCC, hence the huge differences of lesion sizes between HCCs and non-HCCs amongst datasets used for deep learning methods in the literature \cite{wang,yingnature,yasaka}. In this study, we work on a challenging database of \textit{small} hepatocellular carcinomas, which is a private set also used in \cite{ronot}. Average lesion sizes for HCC and non-HCC tumors are reported in Table~\ref{sizes}. While we can note a thin shift of $\sim$4 millimeters between no HCCs and HCCs in $\mathcal{D}_1$, the gap reaches 15 millimeters for $\mathcal{D}_2$, which proves the difficulty of HCC characterization of our training dataset.

\begin{table}[ht]
\centering
\begin{tabular}{c|c|c}
                           & No HCC           & HCC        \\ \hline
\multicolumn{1}{c|}{$\mathcal{D}_1$} & 15.4 ± 5.6  & 19.3 ± 5.6  \\ \hline
\multicolumn{1}{c|}{$\mathcal{D}_2$} & 17.6 ± 12.7 & 32.6 ± 14.8 \\ 
\end{tabular}
\caption{Average lesion diameters for HCCs and no-HCCs in  each dataset. Reported sizes are in \textit{mm}.}
\label{sizes}
\end{table}
\paragraph*{Training and evaluation.} \quad In order to predict the three LI-RADS major features, we train our deep network on $\mathcal{D}_1$ in a stratified 5-fold cross-validation fashion. Each feature corresponds to an independent training. In parallel, we compute our handcrafted features and the lesion size, resulting in a 8-dimensional vector. We then optimize the logistic regression regularization coefficient on $\mathcal{D}_1$, and for each deep architecture (row). Table~\ref{cv} reports the cross-validation results on $\mathcal{D}_1$.\\
We train our models using four different architectures. The first two backbones are inspired by \cite{yasaka} that provide state-of-the art results of 92\% AUC on their database. The first one, called ``Tiny" in Table~\ref{cv} and Table~\ref{test}, is composed of five convolutional layers and two dense layers with 8 channels at the first layer and progressively doubled channels to reach a representation space of 128. Two dense layers are finally added, to reach 2 neurons, with a softmax function to predict probabilities (HCC vs no-HCC). This architecture contains 1.5M of parameters. The second ``Small" architecture is the same architecture with doubled number of channels at each convolutional layer, resulting in a total number of 5.5M of parameters. This architecture is also presented in \cite{sarfatiisbi}. We also provide performances of a ResNet-18 \cite{resnet}, trained from scratch and initialized with weights that were pretrained on 23 different medical imaging datasets \cite{med3d}, as well as pretrained ResNet-50 with frozen first layers for fine-tuning purposes, reducing the number of trainable parameters to 24.9M for ResNet-18 and 29.2M for ResNet-50. We use the AdamW optimizer with a batch size of 32 and trained for 600 epochs, a learning rate of $10^{-5}$ and a weight decay of $10^{-3}$, and for fine-tuning experiments we reduce the learning rate to $10^{-6}$. For data augmentation, we use horizontal and vertical flips, rotations and affine transformations. Training experiments were realized on a Tesla V100 mono-GPU.

We evaluate our method on a private test set, $\mathcal{D}_2$. To do so, we first infer the deep models of each criteria as well as for the HCC on the full set $\mathcal{D}_2$ to obtain predicted probabilities of HCC, APHE, EC and NPW. Then, we apply our handcrafted functions on the base to obtain features of each criteria, which allows us to obtain a feature vector of size 8 with the size feature, as for training. As we regularize with respect to the architecture in line, very small changes were observed in the HF column. For clarity, we report the average HF performance over the regularization used by line.\\
We report results of a cross-validated linear evaluation procedure, \textit{i.e.} we keep the same stratified cross-validation as for $\mathcal{D}_1$, but we replace each validation with the full test set, hence the smaller standard deviations that are reported in Table~\ref{test}. For each backbone, we report the results of choosing only the deep learning features + size (DLF), only the handcrafted features + size (HF), and the concatenation of DLF and HF + size (DLF+HF). Finally, we report the AUC from a non-specialist and a liver-pathology specialist on $\mathcal{D}_1$. The AUC for radiologists is computed by normalizing the LI-RADS score to obtain HCC probability (LR-1,2,3,4,5 are divided by 5 to obtain values between 0 and 1).
\section{Results}
\label{sec:results}
\begin{figure}[H]
    \centering
    \includegraphics[width=0.7\linewidth]{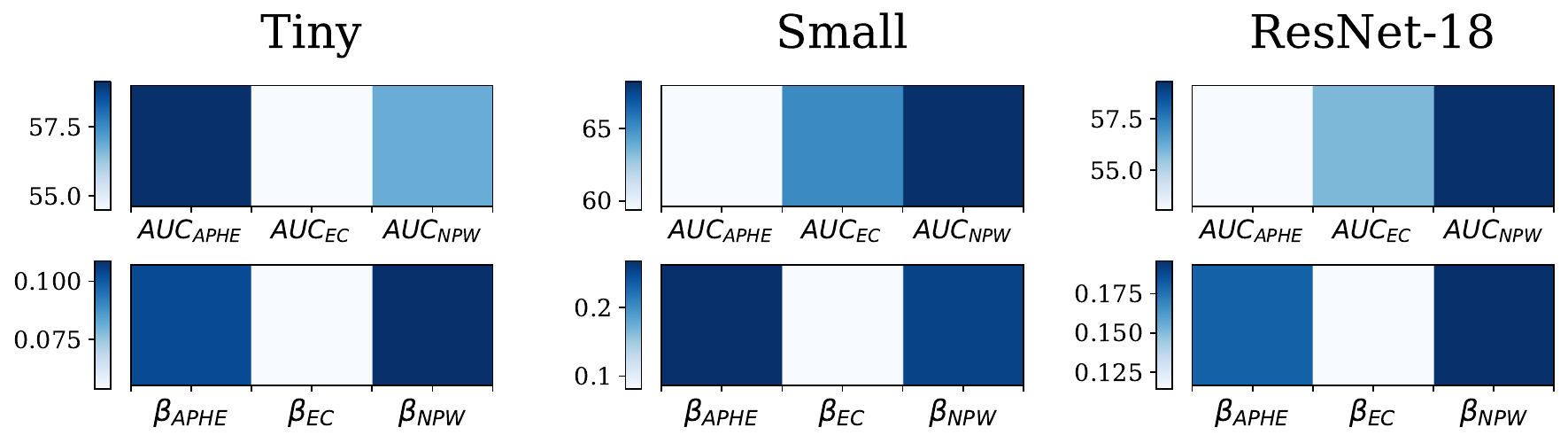}
    \caption{First row: AUC metric for each deep learning model of LI-RADS major features. Second row: absolute values coefficients of the LI-RADS DLF in the logistic regression, for three backbones.}
    \label{coefs}
\end{figure}
In Table~\ref{cv}, we can note that the proposed method outperforms deep learning baselines, with a boost of 6 to 18 points. For $\mathcal{D}_1$, the best AUC is obtained combining our deep and handcrafted features, reaching an AUC of 75\%, which represents a gain of 15 points with respect to a non specialist radiologist, and only 2.7 points away from a liver-pathology specialist. The performance of our method is robust to the backbone used. As a comparison with the state-of-the-art, the authors in \cite{ling} propose a 3D ResNet-18 computed on four-phases CT-scans 3D patches and obtain an AUC of 95.8\%. With the exact same architecture, we obtain a baseline performance of 58.8\% of AUC, which proves the inherent difficulty of our database and the small amount of data available. Pretrained ResNet-18 \cite{med3d} helps however improving the performances, as shown in Table~\ref{cv}. In the same sense, the authors in \cite{yasaka} propose to use a deep learning architecture similar to our small encoder and obtain a groundbreaking AUC of 92\%, while the same architecture obtains 64.4\% on our dataset. It can be noticed that smaller encoders lead to better results in our case, given the small amount of data available.\\
In Table~\ref{test}, we observe that for $\mathcal{D}_2$ the combination of our DLF and HF vectors outperforms both methods taken separately in most of the cases, as well as the deep learning baseline. One can note that performances on this dataset are much higher than on $\mathcal{D}_1$, which reflects the lesion size gap evoked in Table~\ref{sizes}. Our deep learning baseline transfers well enough on this base, proving the robustness of the training.\\
Figure~\ref{coefs} allows visualizing the importance of building an efficient model to predict the LI-RADS major features, before using them to predict HCC. We can indeed draw a parallel between the best individual performance of each predicted feature (by their AUC) and the absolute values of the deep learning major features in the logistic regression for HCC prediction. The logistic regression for HCC naturally selects the most accurately predicted major feature (NPW). However, as we can see with the small encoder performance, a smarter deep features combination could be beneficial for our model performance.

\begin{table*}[htbp]
\centering
\begin{tabular}{c|c|c|c|c|c|c}
\multicolumn{1}{c|}{\textbf{Train → Test}} &  \multicolumn{1}{c|}{\textbf{Model}} & \multicolumn{1}{c|}{\textbf{DL baseline}}       & \multicolumn{1}{c|}{\textbf{DLF}} & \multicolumn{1}{c|}{\textbf{HF}} & \multicolumn{1}{c|}{\textbf{DLF+HF}} & \multicolumn{1}{c}{\textbf{↑ w.r.t baseline}} \\ \hline
\multirow{6}{*}{$\mathcal{D}_1$ → $\mathcal{D}_1$}      & Radiologists         & \multicolumn{4}{c|}{Non specialist: \textit{60.4} / Specialist: \textit{76.7} / AVG: \textit{68.6}}                                                         & \multirow{2}{*}{}                            \\ \hdashline
                                    & Tiny                & \multicolumn{1}{c|}{\textit{68.3 ± 4.3}}             & \multicolumn{1}{c|}{74.2 ± 4.7} & \multicolumn{1}{c|}{\underline{\smash{74.4 ± 3.8}}}          & \textbf{74.9 ± 3.8} & +6.6                                             \\
                                    & Small \cite{yasaka}              & \multicolumn{1}{c|}{\textit{64.4 ± 9.4}}             & \multicolumn{1}{c|}{\underline{\smash{74.7 ± 3.9}}} & \multicolumn{1}{c|}{74.4 ± 3.8}          & {\textbf{75.0 ± 3.8}}
                                    & +10.6\\
                                    & ResNet-18 \cite{ling}             & \multicolumn{1}{c|}{\textit{58.8 ± 4.5}}             & \multicolumn{1}{c|}{\underline{\smash{70.6 ± 2.8}}}         &  \textbf{74.4 ± 3.8}               &  70.5 ± 2.9
                                     & +15.6
                                    \\
                                    & ResNet-18 (Med3D) & \multicolumn{1}{c|}{\textit{59.4 ± 3.7}}             & \multicolumn{1}{c|}{73.7 ± 3.0}
                                    & \textbf{74.4 ± 3.8}
                                    & \underline{\smash{74.2 ± 2.9}}    & +15.0               \\
                                    & ResNet-50 (Med3D) & \multicolumn{1}{c|}{\textit{56.5 ± 3.5}}             & \multicolumn{1}{c|}{71.0 ± 1.9} & \multicolumn{1}{c|}{\textbf{74.4 ± 3.8}}           &        \underline{\smash{{71.2 ± 1.9}}}  &      +17.9                                     \\
                                          
\end{tabular}
\caption{5-fold cross-validation AUCs. Reported average AUCs and std are computed across the five folds. The small and ResNet-18 architectures obtain SOTA results on other datasets, as reported in \cite{ling,yasaka}.}
\label{cv}
\end{table*}

\begin{table*}[h!]
\centering
\begin{tabular}{c|c|c|c|c|c|c}
\multicolumn{1}{c|}{\textbf{Train → Test}} &  \multicolumn{1}{c|}{\textbf{Model}} & \multicolumn{1}{c|}{\textbf{DL baseline}}       & \multicolumn{1}{c|}{\textbf{DLF}} & \multicolumn{1}{c|}{\textbf{HF}} & \multicolumn{1}{c|}{\textbf{DLF+HF}} & \multicolumn{1}{c}{\textbf{↑ w.r.t baseline}} \\ \hline
\multirow{5}{*}{$\mathcal{D}_1$ → $\mathcal{D}_2$}   & Tiny                & \multicolumn{1}{c|}{\textit{77.8 ± 1.8}}             & \multicolumn{1}{c|}{82.2 ± 0.1} & \multicolumn{1}{c|}{\underline{\smash{83.0 ± 0.2}}}          & \textbf{83.2 ± 0.3}    & +5.4                                         \\
& Small \cite{yasaka}                & \multicolumn{1}{c|}{\textit{72.7 ± 4.5}}             & \multicolumn{1}{c|}{82.4 ± 0.2}          & \multicolumn{1}{c|}{\underline{\smash{83.0 ± 0.2}}}           &        \textbf{83.0 ± 0.3}       &    +10.3                                          \\
                                    & ResNet-18 \cite{ling}           & \multicolumn{1}{c|}{\textit{71.0 ± 2.4}}    & \multicolumn{1}{c|}{81.6 ± 0.6}            &  
                                    \textbf{83.0 ± 0.2}& 
                                    \underline{\smash{82.2 ± 0.6}} &
                                    +12.0\\
                                    & ResNet-18 (Med3D) & \multicolumn{1}{c|}{\textit{71.4 ± 2.6}}             & \multicolumn{1}{c|}{82.3 ± 0.1} & \multicolumn{1}{c|}{\underline{\smash{83.0 ± 0.2}}}           &        \textbf{83.2 ± 0.2}  & +11.8\\
                                    & ResNet-50 (Med3D) & \multicolumn{1}{c|}{\textit{63.6 ± 4.3}}             & \multicolumn{1}{c|}{81.7 ± 0.5} & \multicolumn{1}{c|}{\textbf{83.0 ± 0.2}}           &        \underline{\smash{82.4 ± 0.5}}  &   +19.4                                      
\end{tabular}
\caption{5-fold cross-validation of a model trained on $\mathcal{D}_1$ and evaluated on $\mathcal{D}_2$. At each iteration, we use one training fold from $\mathcal{D}_1$ and as validation fold we use the entire test set $\mathcal{D}_2$. We show average and std AUC across the five folds. This explains the lower std compared to Table~\ref{cv}.}
\label{test}
\end{table*}

\section{Discussion and Conclusion}
In this paper, we proposed a novel approach for automatic hepatocellular classification by guiding our training using the radiologists' reading grid, the LI-RADS. We first predict LI-RADS major features using a deep learning approach in order to obtain probability of feature presence. In parallel, we propose a handcrafted approach to obtain relevant statistics related to each major feature. We then combine both approaches in a simple manner to predict HCC probability, and we achieve between 3 and 20 points improvements with respect to deep learning baselines, regardless of the backbone used, even with backbones that achieved state-of-the-art results for HCC classification in the literature.\\
For future work, an interesting perspective would be to improve the combination between DLF and HF, as proposed for instance in \cite{vetil} in which the authors minimize the mutual information between both vectors during pretraining in order to reduce redundancy between DL and HC features. Moreover, in this study we propose a two-step approach for HCC classification: first we extract features and then we train a logistic regression for HCC prediction. A potential idea would be to combine these blocks in an end-to-end approach, \textit{i.e.} integrating the HCC classification as a dense layer after the major features classification layer. This method would involve using a binary cross-entropy computed on the major features (multilabel problem) and the HCC. Pretraining mehods could also be explored, notably using the radiological features to guide the learning of a relevant latent space \cite{sarfatiisbi,sarfatimiccai,ruppli}. Eventually, there is a need for new datasets in HCC classification, notably public datasets (the only existing one exclusively contains HCC cases \cite{tcia_dataset}), in order to test and compare the proposed method.

%\section{Acknowledgments}
%\label{sec:acknowledgments}
\newpage

\paragraph*{Compliance with ethical standards.} This research study was conducted retrospectively using human data collected from various medical centers, whose Ethics Committees granted their approval. Data was de-identified and processed according to all applicable privacy laws and the Declaration of Helsinki.
\paragraph*{Acknowledgments.} This work was supported by Région Île-de-France (ChoTherIA project), ANRT (CIFRE \#2021/1735) and was granted access to the HPC resources of IDRIS under the allocation 2023-A0150314655 and 2023-AD011014090 made by GENCI.
%ES, AB and MR are employed by Guerbet.


\bibliographystyle{IEEEbib}
\bibliography{strings,refs}

\end{document}